\documentclass[letterpaper]{article} 
\usepackage{aaai24}  
\usepackage{times}  
\usepackage{helvet}  
\usepackage{courier}  
\usepackage[hyphens]{url}  
\usepackage{graphicx} 
\usepackage{natbib}  
\usepackage{caption} 
\usepackage{lipsum}
\usepackage{graphicx}
\usepackage{algorithm}
\usepackage{algorithmic}
\usepackage{amsmath}
\usepackage{amssymb}
\usepackage{color}

\usepackage{newfloat}
\usepackage{listings}
\DeclareCaptionStyle{ruled}{labelfont=normalfont,labelsep=colon,strut=off} 
\floatstyle{ruled}
\newfloat{listing}{tb}{lst}{}
\floatname{listing}{Listing}
\title{ArtBank: Artistic Style Transfer with Pre-trained Diffusion Model and Implicit Style Prompt Bank}
\author{
	    Zhanjie Zhang\equalcontrib,
	    Quanwei Zhang\equalcontrib,
	    Guangyuan Li,
	    Wei Xing$^\dagger$,
	    Lei Zhao$^\dagger$,
	    Jiakai Sun,
	    Zehua Lan,
	    Junsheng Luan,
	    Yiling Huang,
	    Huaizhong Lin$^\dagger$,
}
\affiliations{
    Intelligent Vision Lab, Zhejiang University\\
    \{cszzj,cszqw,cslgy,wxing,cszhl,csjk,zjucslzh,l.junsheng121,linhz\}@zju.edu.cn, huangyiling@hotmail.com

}

\usepackage{bibentry}
\usepackage{url}

\begin{document}


\twocolumn[{%
	\renewcommand\twocolumn[1][]{#1}%
	\maketitle
	\begin{center}
		\centering
		\includegraphics[width=0.99\linewidth]{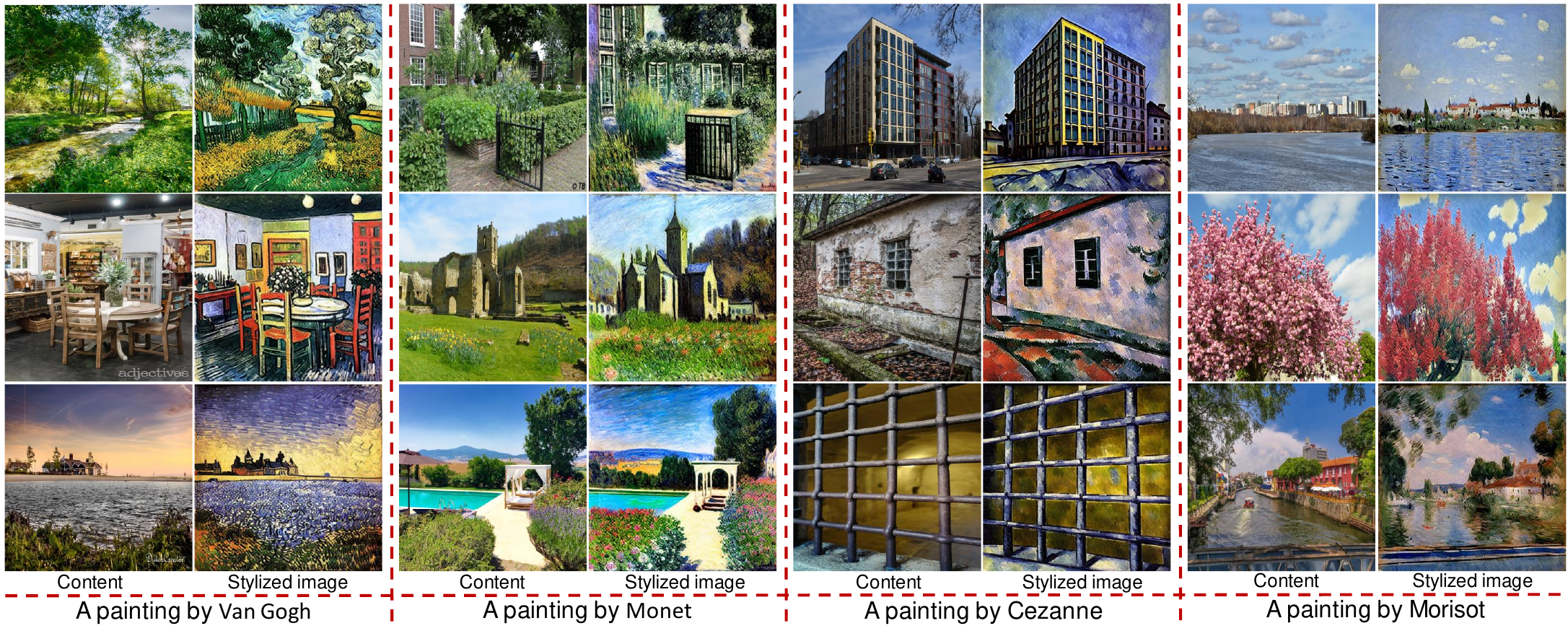}
		\captionof{figure}{Stylized images generated by our proposed ArtBank. With a simple text prompt and a content image, ArtBank can generate highly realistic stylized images with preserving the structure of original content image.}
		\label{teaser}
	\end{center} 
}]

\renewcommand{\thefootnote}{\fnsymbol{footnote}}
\setcounter{footnote}{-1}
\footnote{$^{*}$ Both authors contributed equally to this research.  
	
\hspace{0.185cm}  $^{\dagger}$ Corresponding authors.

\hspace{0.185cm} Copyright \copyright \hspace{0.01cm} 2024, Association for the Advancement of Artificial Intelligence (www.aaai.org). All rights reserved.}

\begin{abstract}
	Artistic style transfer aims to repaint the content image with the learned artistic style. 
	Existing artistic style transfer methods can be divided into two categories: small model-based approaches and pre-trained large-scale model-based approaches. Small model-based approaches can preserve the content strucuture, but fail to produce highly realistic stylized images and introduce artifacts and disharmonious patterns; Pre-trained large-scale model-based approaches can generate highly realistic stylized images but struggle with preserving the content structure. To address the above issues, we propose \textbf{ArtBank}, a novel artistic style transfer framework, to generate highly realistic stylized images while preserving the content structure of the content images. 
	Specifically, to sufficiently dig out the knowledge embedded in pre-trained large-scale models, an \textbf{I}mplicit \textbf{S}tyle \textbf{P}rompt \textbf{B}ank (ISPB), a set of trainable parameter matrices, is designed to learn and store knowledge from the collection of artworks and behave as a visual prompt to guide pre-trained large-scale models to generate highly realistic stylized images while preserving content structure. Besides, to accelerate training the above ISPB, we propose a novel \textbf{S}patial-\textbf{S}tatistical-based self-\textbf{A}ttention \textbf{M}odule (SSAM). The qualitative and quantitative experiments demonstrate the superiority of our proposed method over state-of-the-art artistic style transfer methods.
	  Code is available at \url{https://github.com/Jamie-Cheung/ArtBank}.
\end{abstract}


\section{Introduction}
Artistic style transfer aims to transfer the learned styles onto arbitrary content images to create a new artistic
image. 
Existing artistic style transfer methods can be classified into small model-based methods and pre-trained large-scale model-based methods.

\begin{figure}[htb]
	\centering
	\includegraphics[width=1\columnwidth]{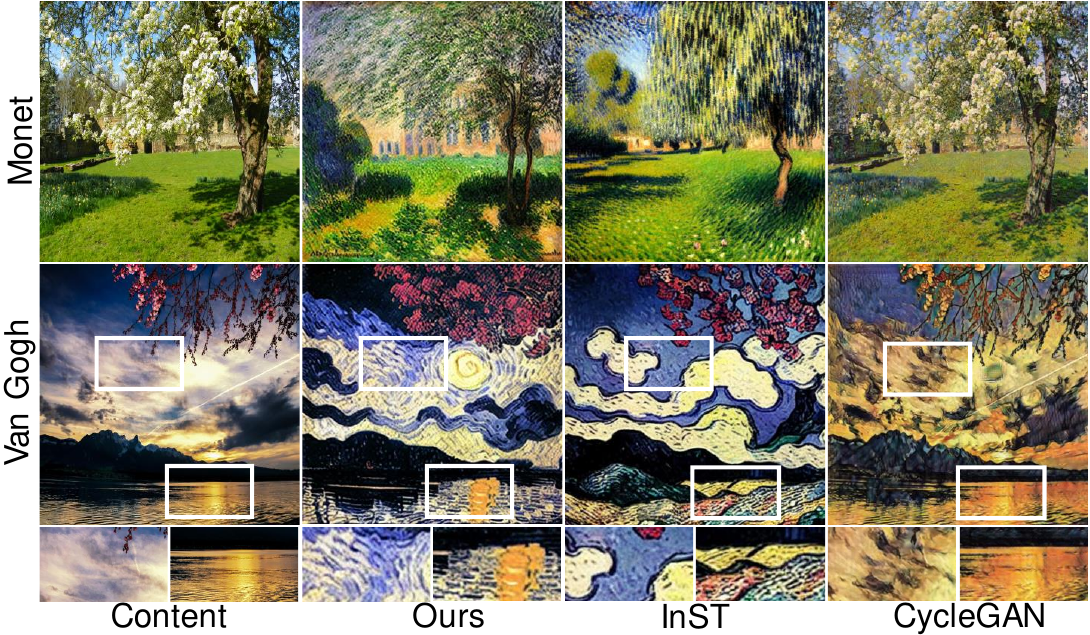}
	\caption{Stylized examples. (a) The $1^{st}$ column shows the content image. The $2^{nd}$ column shows the stylized image by our method. The other two columns show the stylized images produced by the pre-trained large-scale models (e.g., InST~\cite{zhang2022inversion}) and small model-based methods (e.g., CycleGAN~\cite{zhu2017unpaired}).}
	\label{img2}
\end{figure}


More specifically, small model-based methods~\cite{zhang2021generating,sanakoyeu2018style,kim2019u,park2020contrastive,wang2022aesust,zhang2021generating,ijcai2023p157,yang2022gating,zhang2023caster,chen2023testnerf,zuo2023generative,zhao2020uctgan,chen2021artistic,chen2021dualast,chen2021diverse} focus on training a well-designed forward network to learn style information from the collection of artworks. 
To train such forward networks, Zhu et al.~\cite{zhu2017unpaired} first employed a cycle consistency loss to realize the mapping between the style domain and the content domain in the RGB pixel space. AST~\cite{sanakoyeu2018style} proposed a style-aware content loss to learn style from the collection of artworks for real-time and high-resolution artistic style transfer. GcGAN~\cite{fu2019geometry} designed a predefined geometric transformation to ensure that the stylized image maintains geometric consistency with the input content image. CUT~\cite{park2020contrastive} used contrastive learning to push the stylized patch to appear closer to its corresponding input content patch and keep a better content structure. Based on CUT, LseSim~\cite{zheng2021spatially} introduced a more general-spatially-correlative map in contrastive learning which encourages homologous structures to be closer. ITTR~\cite{zheng2022ittr} utilized a transformer-based architecture to capture contextual information from different perceptions locally and globally. 
While these methods are capable of learning style information from a collection of artworks and preserving the content structure, they fail to generate highly realistic stylized images, and introduce disharmonious patterns and evident artifacts (e.g., $4^{th}$ col in Fig.~ \ref{img2}).


The pre-trained large-scale model-based methods can generate highly realistic image since they are trained on large amounts of data and possess large-scale neural network parameters, which opens up the possibility to generate highly realistic stylized images. Recently, some methods~\cite{dhariwal2021diffusion,huang2022draw,nichol2021glide,wu2022creative,ho2020denoising,xie2023learning,xie2023boxdiff} utilized a text prompt to synthesize highly realistic artistic images based on pre-trained large-scale model. The most representative method is Stable Diffusion (SD)~\cite{rombach2022high}, which uses text prompts as guidance to generate stylized images. However, they struggle with preseving content structure. To this end, Ge et al.~\cite{ge2023Rich} proposed to use a rich text editor to solve how to provide a detailed text prompt to constrain content structure; DiffuseIT~\cite{kwon2023diffusion} utilized a pre-trained ViT model~\cite{tumanyan2022splicing} to guide
the generation process of DDPM models~\cite{ho2020denoising} in terms of preserving content structure. Zhang et al.~\cite{zhang2022inversion} proposed a novel example-guided artistic image generation framework called InST related to artistic style
transfer. 
Although these pre-trained large-scale model-based methods can generate highly realistic stylized images and attempt to preserve the content strcutre, they struggle with maintaining the structure of the original content image (e.g., $3^{rd}$ col in Fig.~\ref{img2}).




To address these problems, we focus on how to propose a more effective method that not only can generate highly realistic stylized images but also preserve the structure of the content image. Recently, the pre-trained SD has possessed the massive prior knowledge to generate highly realisc images.
To exploit the prior knowledge in pre-trained SD, we first design a simple text prompt template with the artist's name (e.g., a painting by Van Gogh). Then, we use CLIP~\cite{radford2021learning} to encode the text prompt template and obtain a text embedding space $v_{t}$. Next, we design an Implicit Style Prompt Bank (i.e., multiple learnable parameter matrices) that can learn and store style information from different collections of artworks. Besides, we propose Spatial and Statistical-based Self-Attention Module (SSAM) to project the learnable parameter matrix into the embedding tensor $v_{m}$. Then, $v_{t}$ and $v_{m}$ are concatenated, obtaining condition tensor $c_\theta(y)$. With condition tensor $c_\theta(y)$, our proposed ArtBank can fully use the prior knowledge in pre-trained large-scale models and the knowledge from the collection of artworks, generating highly realistic stylized images with preserving content structure (Please see Fig.~\ref{teaser} and Fig.~\ref{img2}). To demonstrate the effectiveness of proposed ArtBank, we build extensive experiments on different collections of artworks. All the experiments show our method outperforms the state-of-the-art artistic style transfer methods, including small model-based methods and pre-trained large-scale model-based methods. To summarize, our contributions are listed as follows:
\begin{itemize}
	\item We propose a novel framework called ArtBank, which can generate highly realistic stylized images while preserving the content structure. This is realized by the Implicit Style Prompt Bank (ISPB), a set of trainable parameter matrices, which can learn and store the style information of multiple collections of artworks and dig out the prior knowledge of pre-trained large-scale models. 
	\item 
	We propose the Spatial-Statistical Self-Attention Module (SSAM), which focuses on spatial and statistical aspects, to accelerate the training of proposed ISPB.
	\item We have conducted extensive experiments on multiple collections of artworks and synthesized highly realistic stylized images compared to state-of-the-art methods.
\end{itemize}

\section{Related work}
\textbf{Small model-based method.}
The small model-based method refers to training a small-scale forward neural network on a small amount of data. For example, Huang et al.~\cite{huang2017arbitrary} proposed an arbitrary artistic style transfer method that can transfer the style
of a style image onto a content image. Li et al.~\cite{li2017universal} conducted the whitening and coloring transforms (WCT) to endow the content features with the same statistical characteristics as the style features. However, these methods need a reference style image and fail to learn style information from the collection of artworks. To this end, CycleGAN~\cite{zhu2017unpaired}, DiscoGAN~\cite{kim2017learning}, and U-GAT-IT~\cite{kim2019u} adopt generative adversarial networks and a cycle consistency loss to realize the mapping between the style domain and content domain in the RGB pixel space.
These methods can learn style information from the collection of artworks and preserve the structure of the content image, but the cycle consistency loss adds an extra computational burden. Some researchers~\cite{sanakoyeu2018style,kim2019u,park2020contrastive} proposed to leverage the geometry consistency to preserve the structure of the content image.
Although the aforementioned small model-based methods can generate stylized images with preserving the content structure, they fail to synthesize highly realistic stylized images.

\begin{figure*}[htb]
	\centering
	\includegraphics[width=2.1\columnwidth]{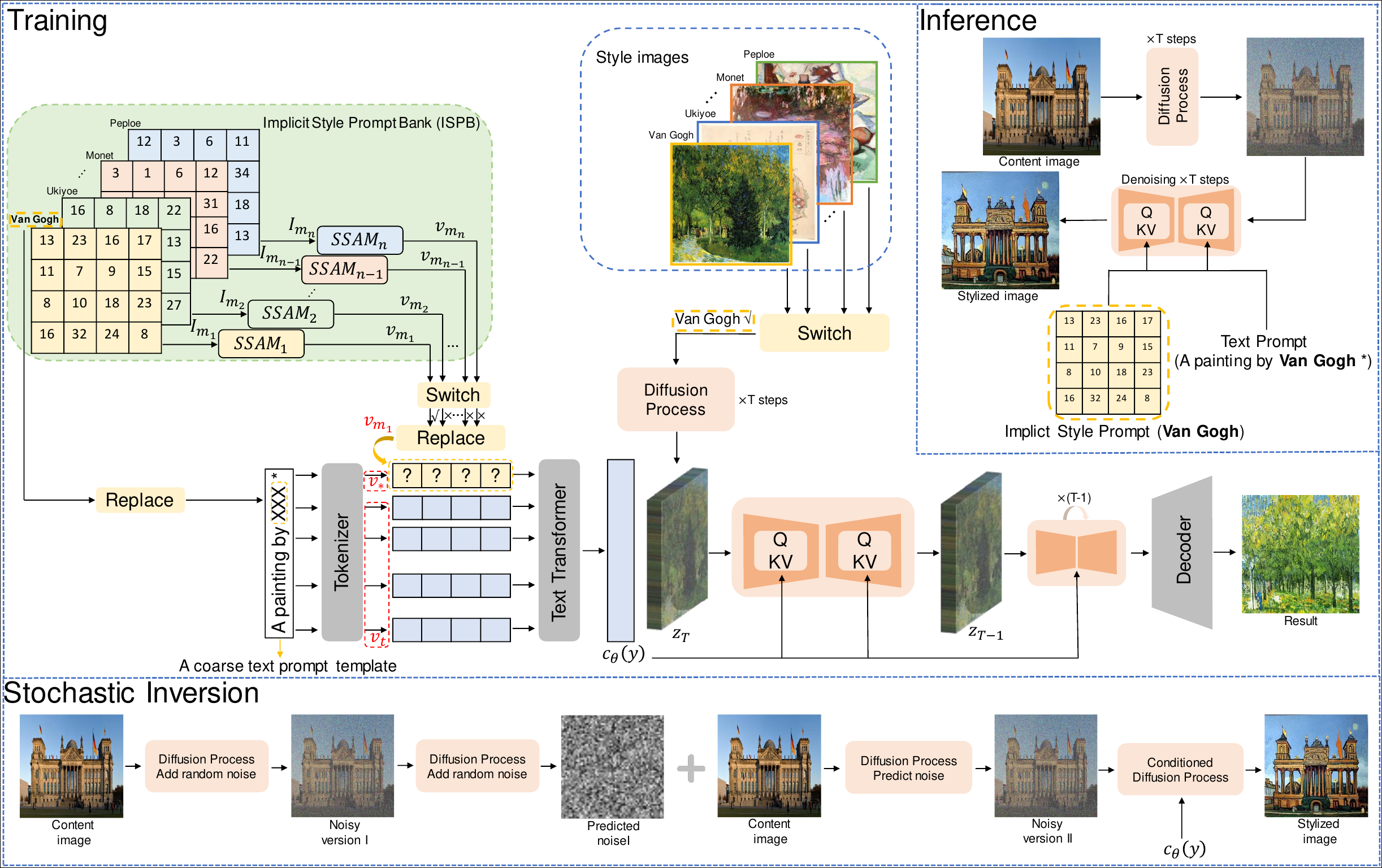}
	\caption{The overview of our proposed ArtBank which consists of two parts: an un-trainable module and a trainable module. The untrainable module is a pre-trained large-scale diffusion model (the model used in this paper is SD 1.4), which can generate highly realistic image. The trainable module is Implicit Style Prompt Bank (ISPB) which can learn and store style information from the collection of artworks. The stochastic inversion (bottom) is used in the diffusion process of inference stage (upper right).
	}
	\label{architecture}
\end{figure*}

\textbf{Pre-trained large-scale model-based method.}
Large-scale models are trained on large amounts of data and can generate highly realistic images. For example, Stable Diffusion~\cite{rombach2022high} is a large-scale text-image generation model which can generate a new highly realistic image corresponding to a text prompt. Pix2pix-zero~\cite{parmar2023zero} first proposed to automatically discover editing directions that reflect desired edits in the text embedding space, and condition diffusion model to generate expired image. Ramesh et al.~\cite{ramesh2022hierarchical} solve the problem of text-conditional image generation by inverted CLIP text embeddings. 
Zhang et al.~\cite{zhang2022inversion} proposed an inversion-based artistic style transfer method to learn the corresponding textual embedding from a single image and use it as a condition to guide the generation of artistic images. DiffuseIT~\cite{kwon2023diffusion} utilized a pre-trained ViT model to guide
the generation process of DDPM models~\cite{ho2020denoising} in terms of preserving content structure. Yang et al.~\cite{yang2023zero} proposed a zero-shot contrstive loss for diffusion models that doesn't require additional fine-tuning or auxiliary networks. 
These methods can perform artistic style transfer from accurate text description or examplar style image but fail to learn and store style information from the collection of artworks. Unlike these methods, our proposed approach learns style information from the collection of artworks based on the proposed ISPB. Our proposed approach does not require explicit text or images as a condition (See in Fig.~\ref{ablation1}) and can synthesize highly realistic artistic images while preserving the structure of the content image. 

\begin{figure*}[htb]
	\centering
	\includegraphics[width=2.1\columnwidth]{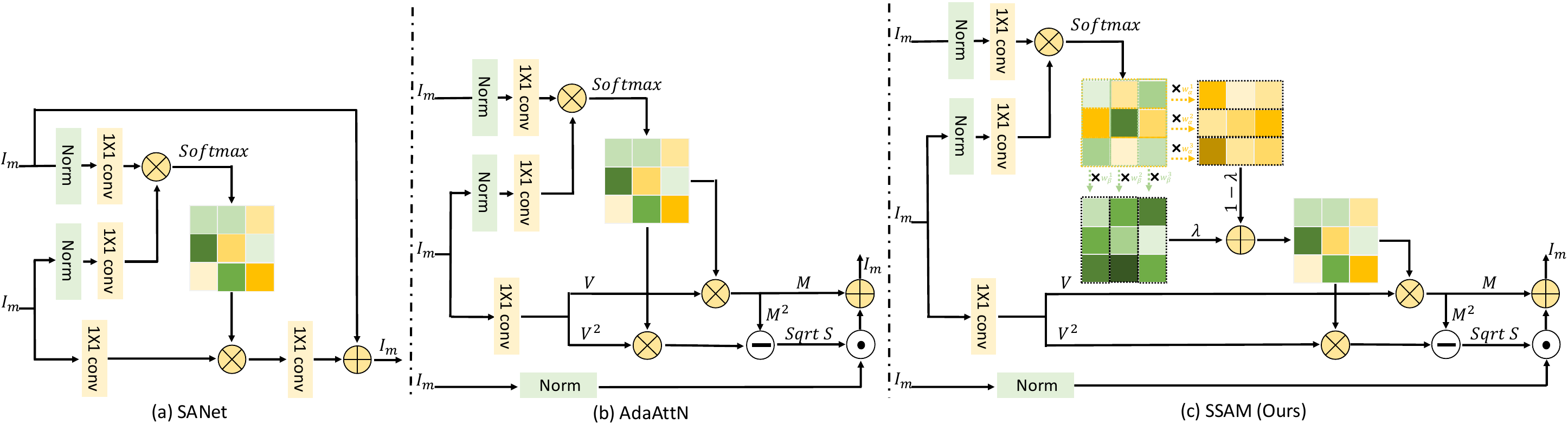}
	\caption{(a) The structure of SANet~\cite{park2019arbitrary}; (b) The structure of AdaAttN~\cite{liu2021adaattn}; (c) The structure of our proposed SSAM. Norm here denotes the mean-variance channel-wise normalization.}
	\label{ssa}
\end{figure*}

%

\section{Method}
\subsection{Overview}
Our proposed ArtBank includes an untrainable part (pre-trained large-scale models) and a trainable part (Implicit Style Prompt Bank, i.e., a set of learnable parameter matrices). The untrainable part utilizes a pre-trained large-scale model (Stable Diffusion, version 1.4) as a backbone, which can generate highly realistic images. The trainable part can learn and store the style information from the collection of artworks and condition pre-trained large-scale model to generate highly realistic stylized images while preserving the structure of the content image. Meanwhile, we propose the Spatial- Statistical Self-Attention Module (SSAM) to accelerate the training of ISPB. Once the training is completed, ArtBank can render arbitrary content images into highly realistic artistic stylized images while preserving the structure of the content image.  

\subsection{Implicit Style Prompt Bank}
In order to learn the knowledge from a collection of artworks, an intuitive way is to unfreeze the parameter of pre-trained large-scale model with the following loss~\cite{nichol2021improved}:
\begin{equation}
\mathcal{L}_{diff}=\mathbb{E}_{z, x, t}\left[\| \epsilon-\epsilon_\theta\left(z_t, t)\right \|_2^2\right],
\end{equation}
where $z \sim E(x), \epsilon \sim \mathcal{N}(0,1)$.
Once the loss function converges, the trained model can render an arbitrary content image into artistic style image.  However, this naive approach will weaken pre-trained large-scale models' ability, learned from previous massive data, to generate highly realistic stylized images.

To dig out massive prior knowledge in pre-trained large-scale model and extract the knowledge from the collection of artworks, we freeze the parameter of pre-trained large-scale model and train an ISPB. ISPB comprises a series of trainable parameter matrices and each trainable parameter corresponds to a collection of artworks. The problem we need to solve is how to teach these learnable parameters to learn and store the style information from the collection of artworks and how to use these trainable parameters to condition the pre-trained large-scale model to generate highly realistic stylized images while preserving content structure. In this paper, we use pre-trained large-scale Stable Diffusion (SD) as backbone. We argue that SD relies on using CLIP-based codes, encoded by CLIP text/image encoder, to guide the denoising process and guide SD to generate desired image. CLIP text encoder can be divided into tokenizer and text transformer modules. If using the text encoder as an example, a text prompt is converted into continuous vector representations $v_{t}$. Although such $v_{t}$ is effective in guiding SD to generate the desired image, it cannot fully dig out the knowledge of SD in style transfer. Based on the above analysis, we first design some coarse text prompt templates (e.g., a painting by Van Gogh *, * is only a meaningless placeholder). The coarse text prompt template is then converted into continuous vector representations: $v_{t}$ and $v_{*}$ (i.e., coarse text prompt is converted into $v_{t}$ and $*$ is converted into $v_{*}$).

In the meantime, the learnable parameter matrix $I_{m}$ of ISPB is also projected into a continuous style representation vector $v_{m}$ by our proposed SSAM (i.e., $v_{m}=SSAM(I_{m})$. The SSAM will be illustrated in Fig.~\ref{ssa}). Then, we replace embedding vector $v_{*}$ with style representation $v_{m}$. Finally, the embedding vectors ($v_{t}$ and $v_{m}$) are transformed into a single conditioning code $c_\theta(y)$. In the above process, only $I_{m}$ and $SSAM$ need to be trained, and each collection of artworks need the corresponding $I_{m_{n}}$ and $SSAM_{n}$. The $SSAM_{n}$ is primarily responsible for accelerating training $I_{m_{n}}$. We use the following loss function for training. 
\begin{equation}
\mathcal{L}_{diff}= \mathbb{E}_{z, x, y, t}\left[\| \epsilon-\epsilon_\theta\left(z_t, t, \operatorname{SSAM}\left(I_{m}), v_{t} \right)\right \|_2^2\right],
\end{equation}
where $z \sim E(x), \epsilon \sim \mathcal{N}(0,1)$ and $v_{t}$ denotes text prompt. Once the $I_{m}$ and $SSAM$ are trained, our proposed ArtBank supports arbitrary content images to generate highly realistic stylized images while preserving content structure.

\subsection{Spatial-Statistical-Wise Self-Attention Module}
\label{ssam}
Fig.~\ref{ssa} illustrates the architecture of our proposed Spatial-Statistical Self-Attention Module (SSAM), which differs from previous self-attention approaches~\cite{park2019arbitrary,liu2021adaattn,li2023dudoinet,li2023self,li2023rethinking,cui2022attention}. Our novel SSAM can learn and evaluate the value change of the parameter matrix from both spatial and statistical perspective. Specifically, we use row-column-wise attention from spatial aspects and mean-variance-wise attention from statistical aspects to extract parameter information. This approach can accelerate the convergence rate, reduce the volatility of parameter matrix updates, and dig out knowledge in SD. The SSAM starts with a trainable parameter matrix $I_{m}$, which is encoded into a query ($Q$), key ($K$), and value ($V$).
\begin{equation}
Q=W_Q \cdot I_{m}, K=W_K \cdot I_{m}, V=W_V \cdot I_{m}
\end{equation}
where $W_Q$, $W_K$, $W_V$ are learnable convolution layer. The attention map $A$ can be calculated as:
\begin{equation}
A=Softmax\left(Q^{\top} \otimes K\right)
\end{equation}
where $\otimes$ denotes matrix multiplication

For attention map $A$, we build col-wise weight matrix $W_{col} \in R^{H_c W_c \times 1}$ and row-wise weight matrix $W_{row} \in R^{1\times  H_c W_c }$. And to make it easier to calculate, we replicate $W_{col}$ and $W_{row}$ along
with col and row, respectively. Then we can obtain col-wise and row-wise attention maps as below.
\begin{equation}
A_{\text {col }}=A \cdot W_{\text {col }}, A_{\text {row }}=A \cdot W_{\text {row }}
\end{equation}
Then, $\hat{A}=\alpha \cdot A_{\text {col }}+(1-\alpha) \cdot A_{\text {row }}$(i.e., $\alpha$ is learnable weight.). Furthermore, 
\begin{equation}
\hat{M}=V\cdot\hat{A}
\end{equation}
The attention-weighted standard deviation $\hat{S} \in R^{C \times H_c W_c}$ as:
\begin{equation}
\hat{S}=\sqrt{(V \cdot V) \otimes \hat{A}^{\top}-\hat{M} \cdot \hat{M}}
\end{equation}
where $\cdot$ represents the element-wise product. Finally, corresponding
scale $\hat{S}$ and shift $\hat{M}$ are used to generate a transformed parameter matrix as:
\begin{equation}
v_{m}=\hat{S} \cdot \operatorname{Norm}\left(I_{m}\right)+\hat{M}
\end{equation}
we redefine this process as: $v_{m} = SSAM(I_{m})$. 

\begin{figure*}[htb]
	\centering
	\includegraphics[width=2.1\columnwidth]{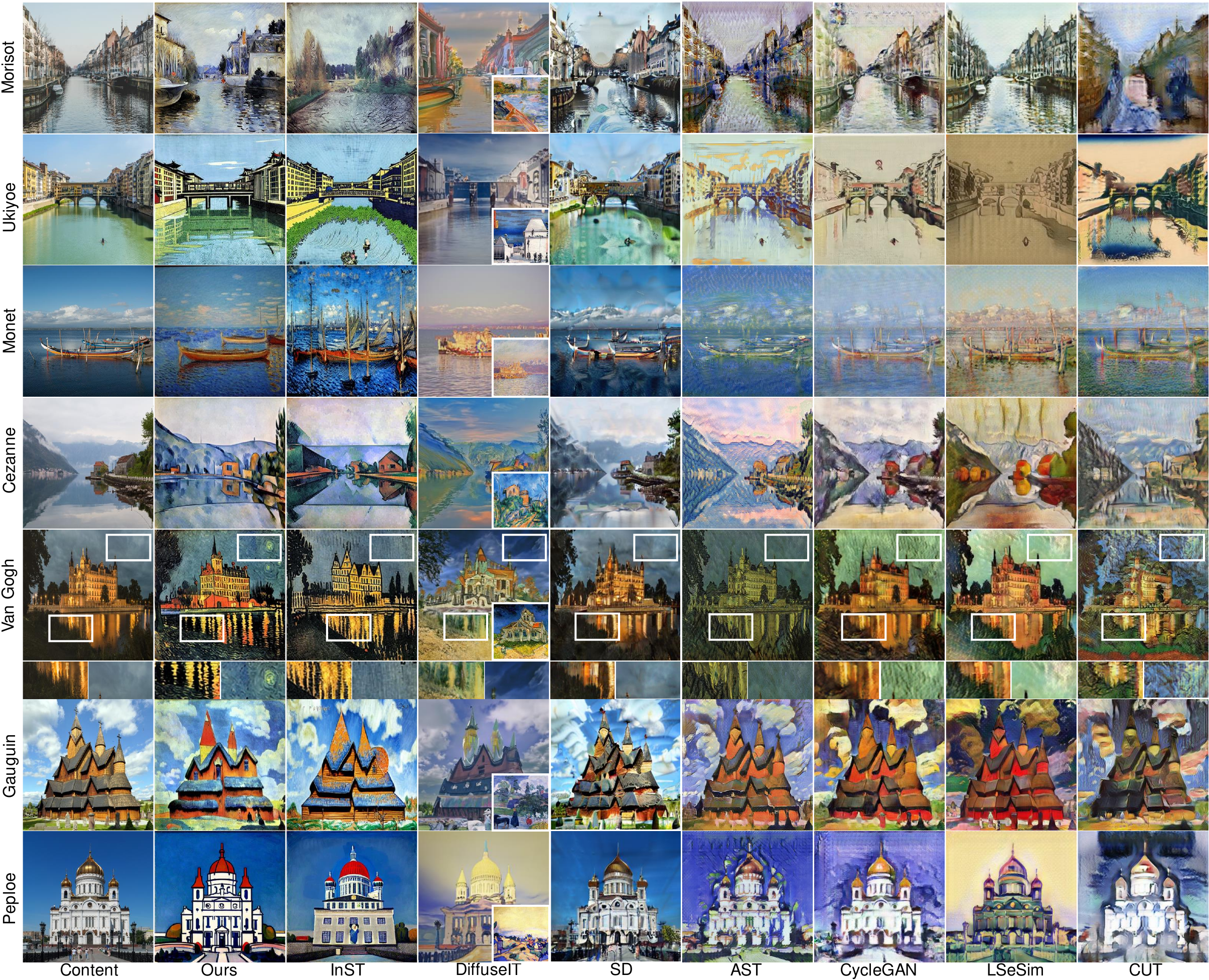}
	\caption{Qualitative comparisons with SOTA artistic style transfer methods.}
	\label{comparison}
\end{figure*}

\subsection{Stochastic Inversion}
In pre-trained large-scale model, random noise plays a crucial role in preserving the content structure of stylized images~\cite{hertz2022prompt}. 
However, random noise is hard to predict, and incorrectly predicted noise can cause a content mismatch between the stylized image and the content image. To this end, we first add random noise to the content image and use the denoising U-Net in
the diffusion model to predict the noise in the image. The
predicted noise is used as the initial input noise during inference to preserve content structure (called stochastic inversion, as shown in bottom of Fig.~\ref{architecture}). Based on this strategy, our proposed ArtBank can generate highly realistic stylized images while preserving better content structure.

\begin{table*}[htb]
	\small
	\caption{Quantitative comparisons with state-of-the-art methods. * denotes the average user preference.}
	\centering
	\setlength{\tabcolsep}{0.1cm}
	\begin{center}		
		\begin{tabular}{c|cccccccccc}
			\hline
			\footnotesize &Ground truth &Ours&InST&SD&DiffuseIT& AST & CycleGAN &LSeSim &CUT 
			\\
			\hline
			\footnotesize Van Gogh&0.7588&\textbf{0.7321}&0.7244&0.5440&0.6632&0.6736&0.6875&0.6727&0.7124
			\\
			\footnotesize Morisot&0.8024&\textbf{0.7447}&0.6983&0.5013&0.6871&0.6659&0.7063&0.6730&0.7389
			\\
			\footnotesize Ukiyoe &0.7495&\textbf{0.7384}&0.7272&0.5235&0.6953&0.6546&0.6504&0.6403&0.6553
			\\
			\footnotesize Monet &0.7910&\textbf{0.7556}&0.7319&0.5031&0.6984&0.7249&0.7351&0.7125&0.7266
			\\
			\footnotesize Cezanne &0.7760&\textbf{0.7646}&0.7332&0.5440&0.7216&0.7143&0.7363&0.7250&0.7563
			\\
			\footnotesize Gauguin &0.8248&\textbf{0.8190}&0.7875&0.6231&0.7528&0.6839&0.7139&0.6730&0.7362
			\\
			\footnotesize Peploe &0.7475&\textbf{0.7355}&0.7032&0.5396&0.6807&0.6677&0.6846&0.6327&0.6982
			\\
			\hline 
			\footnotesize Time/sec&-&3.5725&4.0485&3.7547&32.352&0.0312&0.0312&0.0365&0.0312
			\\
			\footnotesize Preference&-&0.679 *&\textbf{0.572}/0.428&\textbf{0.708}/0.292&\textbf{0.664}/0.336&\textbf{0.685}/0.315&\textbf{0.683}/0.317&\textbf{0.672}/0.328&\textbf{0.769}/0.231
			\\
			\hline 
		\end{tabular}
	\end{center}
	\label{CLIPscore}
\end{table*}
\section{Experiments}

\begin{figure*}[htp]
	\centering
	\includegraphics[width=2.1\columnwidth]{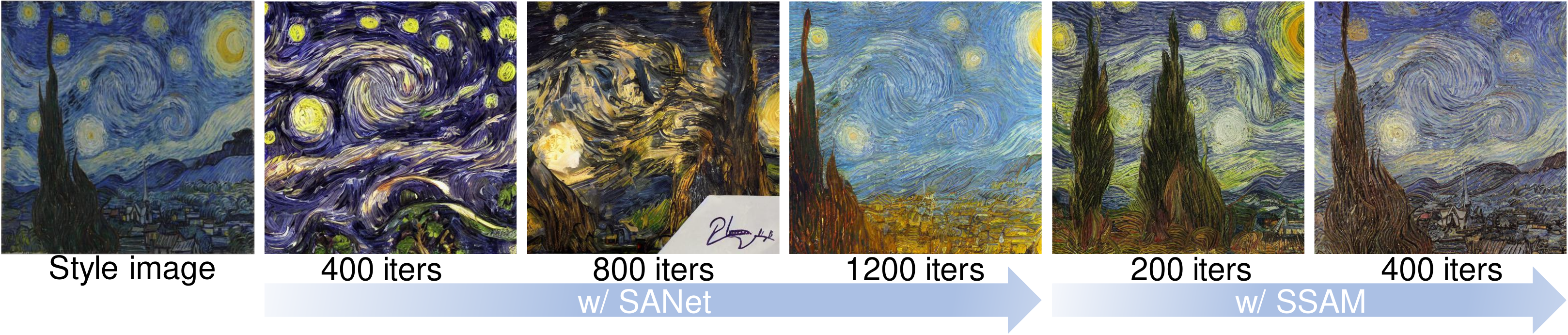}
	\caption{The optimization effiency comparison with SANet~\cite{park2019arbitrary} and our proposed SSAM. }
	\label{optimization}
\end{figure*}
\begin{table}[htb]
	\caption{The CLIP score between text prompt and stylized images.}
	\centering
	\small
	\setlength{\tabcolsep}{0.1cm}
	\begin{center}		
		\begin{tabular}{c|cccccccccc}
			\hline
			\footnotesize &Full Model &w/o Text& w/ SANet& w/ AdaAttN
			\\
			\hline
			\footnotesize Van Gogh&\textbf{0.7321}&0.7248&0.7267&0.7310
			\\
			\footnotesize Morisot&\textbf{0.7447}&0.7383&0.7395&0.7425
			\\
			\footnotesize Ukiyoe &\textbf{0.7384}&0.7256&0.7286&0.7288
			\\
			\footnotesize Monet &\textbf{0.7556}&0.7419&0.7456&0.7462
			\\
			\footnotesize Cezanne &\textbf{0.7646}&0.7532&0.7569&0.7572
			\\
			\footnotesize Gauguin &\textbf{0.8190}&0.8025&0.8036&0.8139
			\\
			\footnotesize Peploe &\textbf{0.7355}&0.7232&0.7285&0.7325
			\\
			\hline 
		\end{tabular}
	\end{center}
	\label{ablation2}
\end{table}

\subsection{Implementation deatails}
We use pre-trained large-scale diffusion model (SD version 1.4) as our backbone. We train our proposed module for each collection of artworks using two NVIDIA GeForce RTX3090 GPUs. The training process requires about 200,000 iterations with a batch size of 1 and takes about two days to complete for each collection. We use a base learning rate of 0.001. The art images are chosen from the Wikiart~\cite{wikiart} dataset and scaled to 512$\times$512 pixels. The training set size varies for each class: 401 for Van Gogh, 130 for Morisot, 1433 for Ukiyoe, 1072 for Monet, 584 for Cezanne, 292 for Gauguin, and 204 for Peploe. During inference, we randomly select some content images from DIV2K~\cite{agustsson2017ntire} as the initial input images.
\subsection{Qualitative Comparisons.} 
\textbf{Comparison with SOTA style transfer methods.}
We compare our method with the sate-of-the-art artistic style transfer methods, including InST~\cite{zhang2022inversion}, DiffuseIT~\cite{kwon2023diffusion},
SD~\cite{rombach2022high}, AST~\cite{sanakoyeu2018style}, CycleGAN~\cite{zhu2017unpaired}, LSeSim~\cite{gao2022learning} and CUT~\cite{park2020contrastive}. 
%
As the representative of small model-based methods,  AST, CycleGAN, LSeSim and CUT can generate stylized images with better content structure; they also introduce artifacts and disharmonious patterns into stylized images.
As shown in Fig.~\ref{comparison}, as the representative of a pre-trained large-scale model, the InST is trained based on the diffusion model and can learn style information from a single style image. To make a fair comparison, we retrained InST and used the same collection of artworks and text prompts with our proposed method. In the inference,  InST used the content image as the initial input image, the text prompt and the content image are used as conditional input. Fig.~\ref{comparison} shows that InST still has limiations in preserving content structure compared to our approach.
DiffuseIT and SD have more limitations in preserving content structure. 

Compared to the above methods, our proposed ArtBank can not only fully mine the knowledge in the pre-trained large-scale model but also learn and store the style information from the collection of artworks to generate highly realistic stylized images while preserving better content structure.


\subsection{Quantitative Comparisons.}
To better demonstrate the superiority of our proposed method in artistic style transfer. We also compare our proposed method with other methods in the terms of CLIP Score, Preference Score and Timing Information.

\textbf{CLIP Score}. CLIP~\cite{radford2021learning} is a cross-modal model pre-trained on 400M image-caption pairs and can be used for robust automatic evaluation of the accuracy between images and text prompt~\cite{hessel2021clipscore}. CLIP Score can measure the similarity between text prompt and the artistic style images. As shown in Tab.~\ref{CLIPscore}, ``Ground Truth" denotes the similarity between text prompt and the collection of artworks. Taking the collection of artworks from Van Gogh as an example, we calculated the mean of similarity between 401 artistic images and a text (a painting by Van Gogh). We also calculate the mean of  similarity between the 1,000 stylized image and a text prompt. 
We employed the same strategy to calculate the CLIP score for the other collection of artworks, such as Morisot, Ukiyoe, Monet, etc. As shown in Tab.~\ref{CLIPscore}, our method achieves a higher CLIP score compared to other state-of-the-art methods, and is even close to the ground truth score.

\textbf{Timing Information}
The $9^{th}$ row of Tab.~\ref{CLIPscore} shows the run time comparisons on images with a scale of 512$\times$512 pixels. Although our proposed method does not have the advantage of inference efficiency compared with the small method-based methods, it is significantly faster than the pre-trained large-scale model-based model.

\textbf{Preference Score}~\cite{chen2021artistic,zhang2022inversion}.To measure the popularity of stylized images generated by two artistic style transfer methods, preference score is commonly used. In this section, we randomly selected 100 content images as input for our proposed method and existing artistic style transfer methods, generating 100 stylized images. To ensure a fair and efficient calculation of the preference score, we asked each participant to select their preferred stylized image one at a time from a set of 10 images generated by our method and 10 images generated by one of the other methods. Participants were instructed to prioritize artistic authenticity and content continuity between stylized images and content images. We recruited 200 participants to repeat the above process, and collected a total of 2,000 votes. Tab.~\ref{CLIPscore} shows the percentage of votes for each artistic style transfer method, indicating that our proposed method was the most popular.
\subsection{Ablation Study}
\begin{figure}
	\includegraphics[width=1\columnwidth]{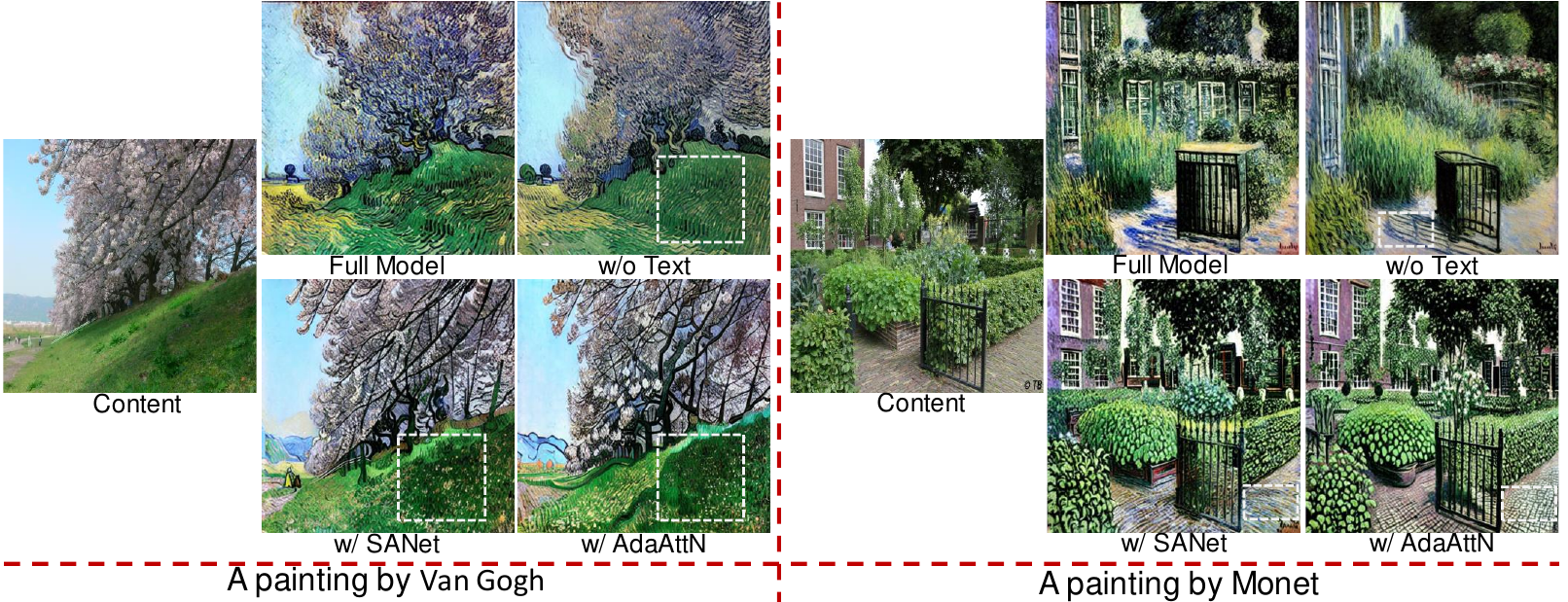}
	\caption{The ablation study of attention module and text prompt. }
	\label{ablation1}
\end{figure}

Self-attention can faster optimization efficiency of diffusion model~\cite{hessel2021clipscore}. To demonstrate that our proposed method with the proposed SSAM can optimize the target style image faster, we show the optimization process (see in in Fig.~\ref{optimization}) using our proposed SSAM or SANet~\cite{park2019arbitrary}. While using SANet requires 1400 iters to achieve incomplete convergence, SSAM only requires 400 iters. Besides, we retrained ISPB using SANet and AdaAttN with the same iterations. As shown in Fig.~\ref{ablation1}, we observe that stylized images with other attention mechanisms are less creative and exhibit fewer brushstroke details. The quality of stylized images also degraded in detail when the text prompt is removed.
To further validate the effectiveness of our proposed module from quantitative evaluation, we also calculate CLIP score, as shown in Tab.~\ref{ablation2}.



\section{Conclusion} We introduce a novel artistic style transfer framework, called as ArtBank, which can addresses the challenge of digging out the knowledge from pre-trained large models to generate highly realistic stylized images while preserving the content structure of the original content images. 
Extensive experiments demonstrate that our proposed method achieves state-of-the-art performance in artistic style transfer compared to existing SOTA methods. In the future, we lookforward to designing a more effective visual prompt to fully dig out the prior knowledge of pre-trained large-scale model in generating highly realisc stylized images.

\section{Acknowledgments}
This work was supported in part by Zhejiang Province Program (2022C01222, 2023C03199, 2023C03201, 2019007, 2021009), the National
Program of China (62172365, 2021YFF0900604, 19ZDA197), Ningbo Program (022Z167), and MOE Frontier Science Center for Brain Science 
\& Brain-Machine Integration (Zhejiang University).

\bibliography{aaai24}

\end{document}